\RequirePackage[T1]{fontenc}
\documentclass[letterpaper,10pt,conference]{ieeeconf}

\usepackage{graphicx}
\usepackage{xcolor}

\usepackage{array}
\usepackage{booktabs}
\usepackage{tabularx}
\usepackage{amsmath}
\usepackage{amssymb}
\usepackage[font=footnotesize]{caption}
\newcommand{\mavptablestyle}{%
  \footnotesize
  \setlength{\tabcolsep}{3pt}%
  \renewcommand{\arraystretch}{1}%
}
\usepackage{microtype}
\usepackage{cite}
\usepackage{xurl}
\usepackage[colorlinks=true,linkcolor=black,citecolor=black,urlcolor=blue]{hyperref}
\usepackage{adjustbox}
\usepackage{flushend}
\newcommand{\firstlead}[1]{\par\noindent\textbf{#1}\enspace\ignorespaces}
\newcommand{\lead}[1]{\par\noindent\textbf{#1}\enspace\ignorespaces}

\usepackage{wrapfig}
\title{\LARGE \bf MAVP: Map-Aware Visuomotor Policies for Mobile Manipulation}
\author{
Jinhe Tang$^{1,\dagger}$, Ruixiao Dai$^{1,\dagger}$, and Weiming Zhi$^{1*}$\\
{\normalsize
$^{1}$School of Computer Science, The University of Sydney, Australia}\\
{\normalsize
$^{\dagger}$These authors contributed equally.}\\
{\normalsize
$^{*}$Corresponding author: \texttt{Weiming.Zhi@sydney.edu.au}}\\
{\normalsize Project website: \url{https://123qwedsa123.github.io/mavp/}}
}

\newcommand{\mavpfigure}{%
\begin{minipage}{\textwidth}
\captionsetup{type=figure,skip=5pt}
\centering
\begingroup
\setlength{\unitlength}{\dimexpr\linewidth/4320\relax}
\begin{picture}(4320,2112)
\put(0,0){\includegraphics[width=\linewidth]{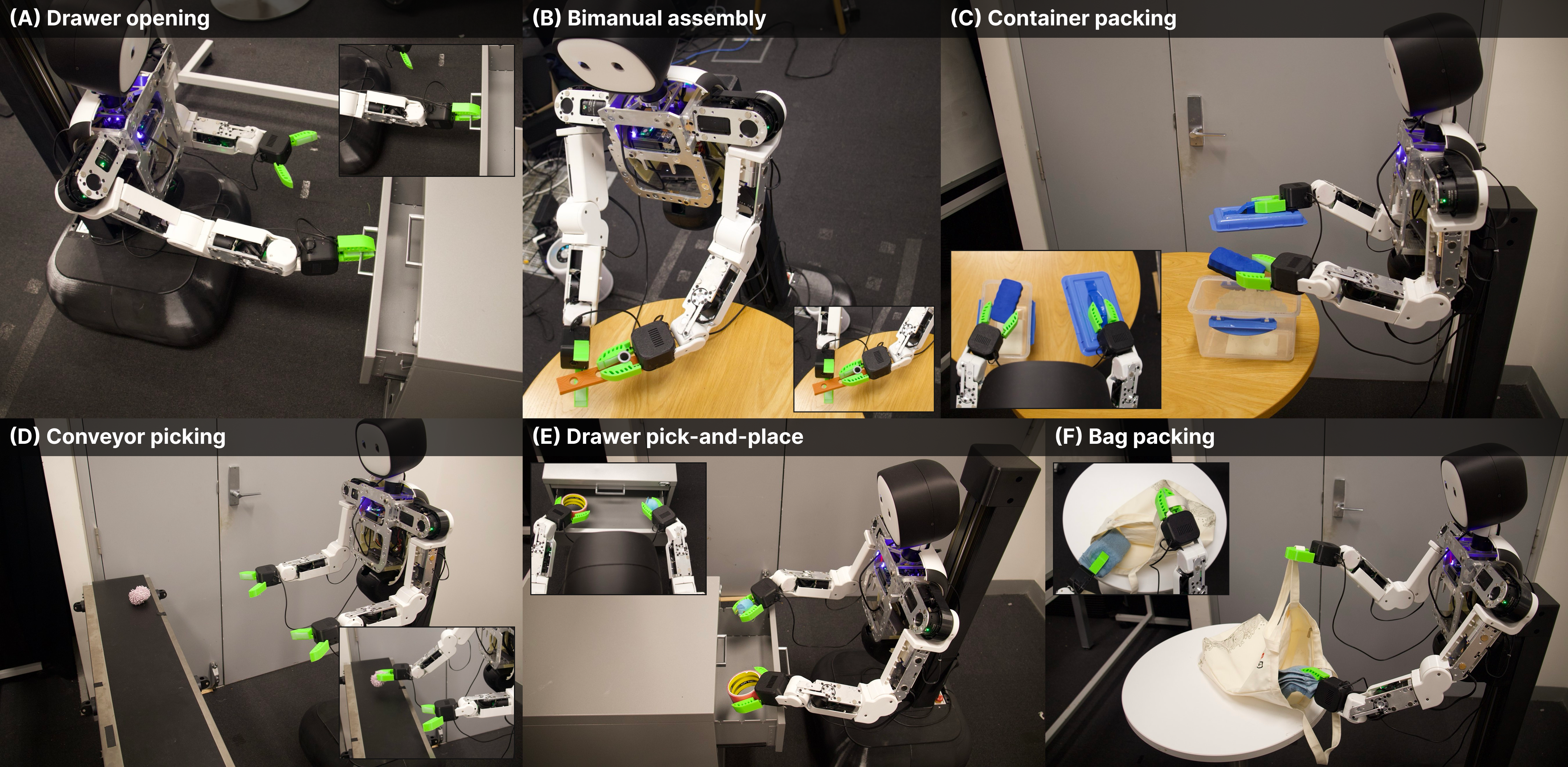}}
\put(0,2004){\color{black}\rule{4320\unitlength}{108\unitlength}}
\put(0,852){\color{black}\rule{4320\unitlength}{108\unitlength}}
\newcommand{\taskfiglabel}[3]{\put(##1,##2){\makebox(0,0)[l]{\color{white}\sffamily\bfseries\fontsize{8}{9}\selectfont ##3}}}
\taskfiglabel{28}{2058}{(A) Drawer Packing}
\taskfiglabel{1468}{2058}{(B) Disassemble and Deliver}
\taskfiglabel{2620}{2058}{(C) Lidded Box Packing}
\taskfiglabel{28}{906}{(D) Conveyor Picking}
\taskfiglabel{1468}{906}{(E) Dual-Drawer Return}
\taskfiglabel{2908}{906}{(F) Bag Packing}
\end{picture}
\endgroup
\caption{We propose Map-Aware Visuomotor Policies (MAVP), evaluated on six real-world mobile manipulation tasks requiring coordinated base motion and manipulation across multiple workspace locations.}

\label{fig:task-overview}
\end{minipage}

}
\IEEEoverridecommandlockouts
\IEEEaftertitletext{\makebox[\textwidth][c]{\mavpfigure}}

\begin{document}\raggedbottom

\maketitle

\thispagestyle{empty}
\pagestyle{empty}

\begin{abstract}
Successful mobile manipulation requires coordinated base and arm motion while maintaining accurate spatial positioning. However, demonstration-trained policies can struggle to realise the intended base motion reliably, leading to spatial misalignment and subsequent manipulation failures. We present MAVP (Map-Aware Visuomotor Policies), a framework that improves execution reliability by predicting explicit base-pose targets and tracking them using localisation feedback. MAVP reconstructs a static map from teleoperated demonstrations and expresses demonstrated base trajectories in a shared map frame, providing consistent spatial supervision across demonstrations. At execution time, the policy receives RGB observations, joint states, and the robot’s current map-frame base pose, and jointly predicts target base poses, arm actions, and gripper actions. A low-level controller tracks the predicted base targets using feedforward motion and pose-error feedback, enabling correction of execution deviations. We additionally use pose-noise augmentation during training to improve robustness to errors in the policy's pose input. Across six real-world manipulation tasks and three policy families, MAVP achieves higher task success rates than unanchored velocity control in all tasks. Videos and additional results are available at \url{https://123qwedsa123.github.io/mavp/}.
\end{abstract}

\begin{figure*}[!t]
\centering
\includegraphics[width=\linewidth]{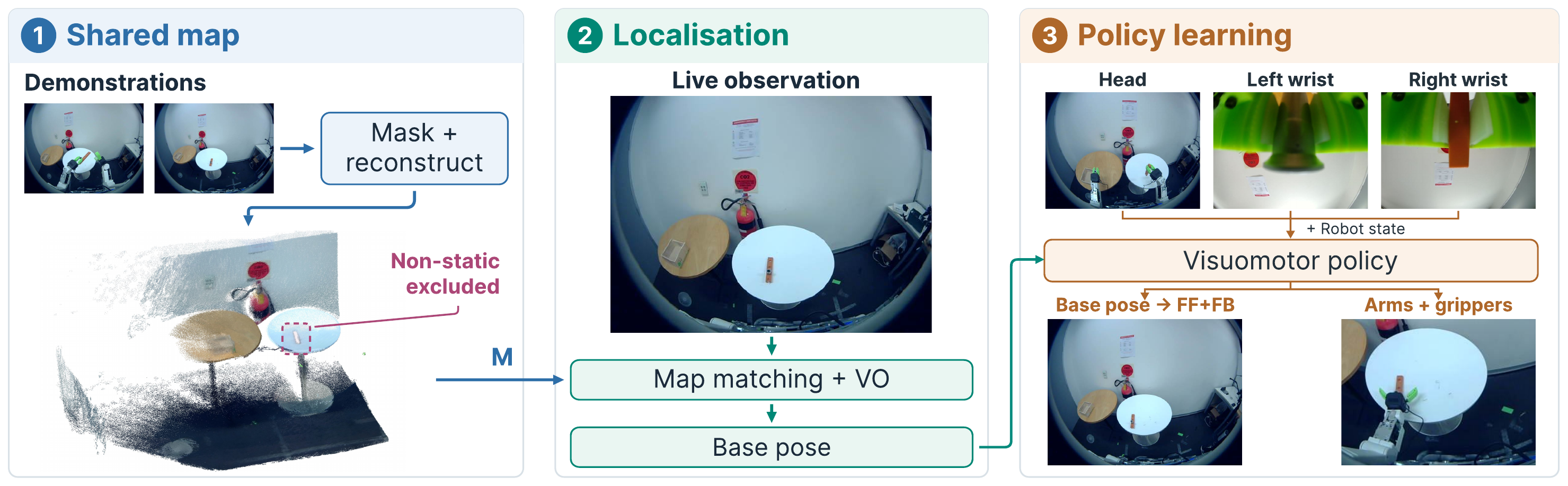}
\caption{Overview of MAVP: map localisation and visual odometry (VO) estimate the base pose in a shared static map reconstructed from masked demonstrations. The policy jointly predicts base-pose targets and arm/gripper actions; feedforward (FF) and pose-error feedback (FB) track the base targets (camera views are illustrative).}
\label{fig:method-overview}
\end{figure*}

\section{Introduction}

Mobile manipulation requires robots to coordinate base and arm motion across multiple locations. For example, disassembling an object and delivering its parts requires the base to reach suitable poses for manipulation at successive locations. Base position and orientation affect both the arms' reachable workspace and the camera viewpoints, so positioning errors can disrupt subsequent manipulation. Learning from demonstration enables robots to acquire motion skills through learned dynamical systems~\cite{zhi2022diffeomorphic,zhi2022inn} or probabilistic task-space trajectory models~\cite{zhi2024rptl}. For mobile manipulation, teleoperated demonstrations support learning coordinated base and arm behaviours~\cite{fu2024mobilealoha}, but reliable execution remains challenging.

Prior work explores representations and coordination for mobile manipulation. Mobi-$\pi$ selects suitable base poses for a pretrained manipulation policy~\cite{yang2025mobipi}, while HoMeR combines learned end-effector pose targets with whole-body control~\cite{sundaresan2026homer}. Spatial Action Maps represent actions in a spatial map~\cite{wu2020spatialactionmaps}, and other policies use features from a learned 3D map~\cite{kim2026sbp}. 

Without an explicit base-pose target, a velocity-based policy controls base motion incrementally, causing small prediction and tracking errors to accumulate into base-pose drift over time. The policy must also infer its spatial state from visual and proprioceptive observations to continuously adjust the base motion. Pose-based targets instead provide an explicit spatial objective for feedback control. However, independently initialised coordinate frames across demonstrations can assign different coordinates to the same base pose in the workspace. A shared map resolves this ambiguity by providing a consistent frame for demonstration supervision and target tracking.

We present \textbf{MAVP} (Map-Aware Visuomotor Policies), which expresses demonstrated base trajectories, current map-frame base-pose estimates, and predicted base targets in a shared map frame. The policy receives visual observations, joint states, and the estimated base pose, and jointly predicts base-pose targets, arm actions, and gripper actions. A low-level controller converts the base targets into velocity commands using feedforward motion and pose-error feedback, allowing execution deviations to be corrected without requiring the policy to predict each corrective velocity command.

MAVP constructs its spatial reference from teleoperated demonstrations, masking robot arms and moving objects before reconstructing a static scene map. During execution, visual localisation, camera calibration, and robot kinematics provide the base-pose estimate. To improve tolerance to imperfect localisation, we perturb the policy's pose inputs during training while retaining the demonstrated action targets.

Our technical contributions include,
\begin{itemize}
    \setlength{\itemsep}{0pt}
    \item \textbf{Shared map construction from demonstrations.} We construct a shared static map from teleoperated demonstrations by masking robot arms and dynamic regions before reconstruction. The map provides a common spatial reference for aligning demonstrations and expressing base states and action targets across episodes.

    \item \textbf{A map-aware policy learning and execution framework.} We introduce MAVP, which conditions the policy on the current map-frame base pose and jointly predicts base-pose targets, arm actions, and gripper actions. A feedforward-plus-feedback controller tracks the base targets, while pose-noise augmentation improves robustness to errors in the policy pose input.

    \item \textbf{Evaluation across tasks, policy families, and system components.} We rigorously evaluate and analyse the properties of MAVP on real-world tasks using different policy classes. 
\end{itemize}

\section{Related Work}\label{sec:related-work}

\firstlead{Mobile manipulation and spatial representations:} Mobile ALOHA learns whole-body actions from teleoperation~\cite{fu2024mobilealoha}. M3 composes reinforcement-learned manipulation and navigation skills~\cite{gu2023m3}, and Mobi-$\pi$ selects base poses for a pretrained manipulation policy~\cite{yang2025mobipi}. UMI-on-Legs and HoMeR connect learned end-effector targets to whole-body controllers~\cite{ha2025umionlegs,sundaresan2026homer}. HoMMI uses a gripper-centred frame for hand-eye policy observations and actions~\cite{xu2026hommi}. Spatial Diagrammatic Instructions express spatial objectives and constraints from sketches for mobile-base placement~\cite{sun2024sdi}. Other methods use point clouds or maps for geometric context~\cite{ze2024dp3,wu2020spatialactionmaps,kim2026sbp}.

MAVP learns base-pose targets in a shared map frame and tracks them with localisation feedback, while jointly predicting arm and gripper actions.

\lead{Action representation and imitation robustness:} Automatic Waypoint Extraction (AWE) compresses demonstrations into spatial waypoints~\cite{shi2023awe}, while HYDRA combines sparse waypoints, dense actions, and action relabelling~\cite{belkhale2023hydra}. Diffeomorphic transforms encode modular robot motions and compose them to adapt demonstrated behaviours to changed surroundings~\cite{zhi2022diffeomorphic}. DAgger addresses policy-induced distribution shift through expert labels on learner-visited states~\cite{ross2011dagger}, and DART perturbs demonstration execution to collect corrective behaviour~\cite{laskey2017dart}. MAVP addresses localisation uncertainty through pose-noise augmentation without collecting corrective demonstrations.

\lead{Mapping and localisation:} Scene mapping and localisation methods reconstruct landmarks, match images, estimate camera poses, and mask dynamic regions~\cite{schoenberger2016sfm,sarlin2019coarse,bescos2018dynaslam,goli2025romo,vggt4d2025}. Learned matchers such as SuperGlue and LightGlue use contextual feature relationships to estimate correspondences~\cite{sarlin2020superglue,lindenberger2023lightglue}. Joint calibration and scene representation methods align reconstructed geometry with robot frames~\cite{zhi2024jcr,tang2026bijcr}. In MAVP, established mapping and localisation techniques provide
a shared scene reference for base-pose supervision and execution feedback.

\section{MAVP: Map-Aware Visuomotor Policies}
\label{sec:method}

MAVP uses a shared map as a common spatial reference for teleoperated demonstrations, policy learning, and closed-loop execution. As shown in Figure~\ref{fig:method-overview}, the shared map is first reconstructed from the demonstrations while excluding non-static scene content. During deployment, live observations are then matched against this map, together with visual odometry, to estimate the current base pose in the map frame. Finally, the estimated pose provides the spatial reference for the visuomotor policy, which predicts map-referenced base targets together with manipulation actions.

\lead{System setup:}
MAVP assumes that the mapped scene geometry remains static. In our setup, we use a dual-arm wheeled robot with calibrated head and wrist cameras. The head-mounted stereo RGB cameras provide map reconstruction and visual odometry; calibrated camera-to-base extrinsics transform camera-pose estimates into map-frame base poses.

\subsection{Shared Map Construction}

\begin{figure}[t]
\centering
\includegraphics[width=\linewidth]{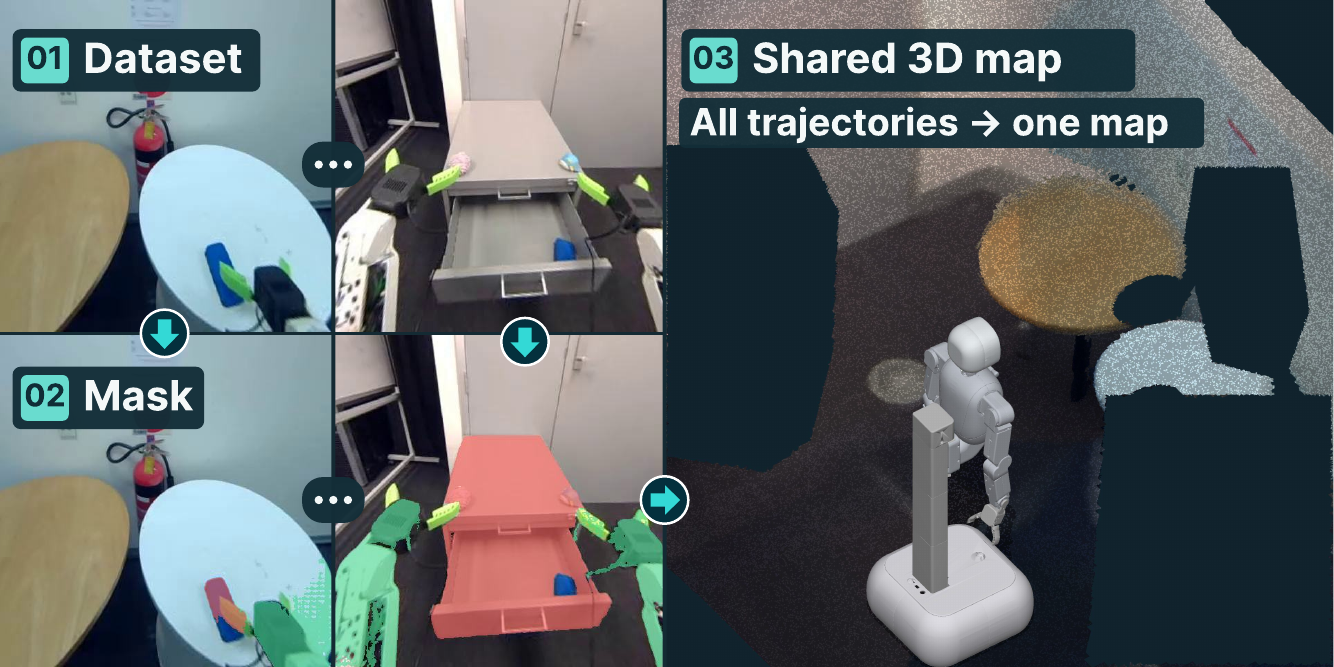}
\caption{Shared map reconstruction from stereo demonstrations after masking robot arms and dynamic regions.}
\label{fig:map-construction}
\end{figure}

MAVP builds a static map offline from demonstrations. Robot-arm and dynamic-object regions are masked out so that the map is reconstructed from static scene features. Figure~\ref{fig:map-construction} summarises the static-scene filtering and shared map reconstruction pipeline, together with the resulting reconstructed map.

\vspace{0.3em}
\noindent\textbf{Reference Sampling and Static-Scene Filtering:}
\label{sec:static-scene-filtering}
MAVP uniformly samples stereo image pairs from the head-mounted cameras in each demonstration to construct the shared map $\mathcal{M}$. Let $I_n^c$ denote the $n$-th sampled image from camera $c\in\{L,R\}$, where $L$ and $R$ denote the left and right stereo cameras. To exclude the robot from map reconstruction, we project the robot model into each image using the joint configuration, forward kinematics, and calibrated camera parameters, yielding a binary robot-arm mask $A_n^c$. Here, $A_n^c(u)=1$ marks pixel $u$ as part of a projected robot arm.

Dynamic scene regions can violate the cross-view consistency required for static map reconstruction. We therefore apply VGGT4D~\cite{vggt4d2025} independently to the image sequence from each head-camera view to separate dynamic elements from the static scene. VGGT4D extracts dynamic cues from VGGT's global attention, aggregates them over a temporal window, and applies projection-gradient refinement to sharpen the resulting region boundaries. This yields a binary dynamic-region mask $D_n^c$ for each sampled image, where $D_n^c(u)=1$ indicates that pixel $u$ is classified as dynamic.

We combine the robot-arm and dynamic-region masks to retain only regions used for static map reconstruction:
\begin{equation}
M_n^c(u)
=
\bigl(1-A_n^c(u)\bigr)
\bigl(1-D_n^c(u)\bigr).
\label{eq:usable-region-mask}
\end{equation}
Thus, $M_n^c(u)=1$ indicates that pixel $u$ is retained for map reconstruction, while robot-arm and dynamic regions are excluded. We denote the sampled images and their corresponding mapping masks by
$\mathcal{I}_{M}=\{(I_n^c,M_n^c)\}_{n,c}$.

\vspace{0.3em}
\noindent\textbf{Shared Map Reconstruction:}
Using the retained regions in $\mathcal{I}_{M}$, MAVP extracts and matches visual features across sampled frames, demonstrations, and stereo views. COLMAP~\cite{schoenberger2016sfm} performs two-view geometric verification to reject inconsistent matches, and then reconstructs a shared sparse map by registering the reference images and triangulating matched features into 3D landmarks. Bundle adjustment jointly refines the camera poses and landmark positions while keeping the calibrated camera intrinsics and stereo geometry fixed. The resulting map $\mathcal{M}$ contains the reference camera poses, 3D landmarks, and their associated image features, which are subsequently used for visual localisation.

\subsection{Map-Based Localisation}

During execution, MAVP localises the head camera in the shared map and transforms the estimated camera pose into a map-frame base pose using the calibrated camera geometry and robot kinematics. The resulting base pose is used by the policy and feedback controller.

\vspace{0.3em}
\noindent\textbf{Visual Query Localisation}
At execution time, MAVP localises the head-camera observations against the shared static map. Features on the robot arm move with the robot and therefore cannot serve as stable references in the map. We apply the robot-arm masking procedure described in Sec.~\ref{sec:static-scene-filtering} to the current stereo images and exclude features within the projected robot regions from localisation.

MAVP first retrieves candidate reference images from the shared map based on visual similarity. SuperPoint~\cite{detone2018superpoint} detects salient local keypoints in both the query and reference images and computes a descriptor for each keypoint. LightGlue~\cite{lindenberger2023lightglue} then compares these keypoints and descriptors across the image pair to identify geometrically consistent feature matches. Because the matched reference features are already associated with 3D landmarks in $\mathcal{M}$, each accepted image match links a 2D query keypoint to a corresponding 3D map point.

Perspective-$n$-Point (PnP) estimation with random sample consensus (RANSAC) recovers the head-camera pose from these 2D--3D matches. Pose estimates must meet inlier-count, reprojection-error, and stereo-consistency thresholds.

\vspace{0.3em}
\noindent\textbf{Map--Odometry Alignment and Base-Pose Estimation:}
MAVP uses cuVSLAM stereo visual odometry~\cite{korovko2025cuvslam} to continuously track the head-camera motion in a local coordinate frame. cuVSLAM recovers metric geometry from stereo correspondences and estimates relative camera motion by tracking features across frames, producing a continuous local pose trajectory. Whenever map-based localisation provides the camera pose in the shared map, we align the odometry trajectory using paired local and map-frame poses.

Let $L$, $M$, $C$, and $B$ denote the local tracking, shared map, head-camera, and mobile-base frames. The rigid transformation ${}^{X}\mathbf{T}_{Y,t}$ maps coordinates from frame $Y$ to frame $X$ at time $t$. cuVSLAM provides ${}^{L}\mathbf{T}_{C,t}$, and map localisation provides ${}^{M}\mathbf{T}_{C,t}$ at the corresponding timestamp.

Using these two corresponding poses, we compute the alignment $\mathbf{T}_{\mathrm{align},t}$ from the local tracking frame to the shared map frame as
\begin{equation}
\mathbf{T}_{\mathrm{align},t}
=
{}^{M}\mathbf{T}_{C,t}
\left({}^{L}\mathbf{T}_{C,t}\right)^{-1}.
\label{eq:map-odom-alignment}
\end{equation}
Applying $\mathbf{T}_{\mathrm{align},t}$ transforms the continuous cuVSLAM trajectory into the shared map frame. Subsequent map localisations update this alignment, with smoothing to limit pose jumps.

The mobile-base pose is then obtained from the aligned camera pose using the camera-to-base transformation:
\begin{equation}
{}^{M}\mathbf{T}_{B,t}
=
\mathbf{T}^{\mathrm{smooth}}_{\mathrm{align},t}
{}^{L}\mathbf{T}_{C,t}
\left({}^{B}\mathbf{T}_{C,t}\right)^{-1}.
\label{eq:map-base-pose}
\end{equation}
Here, $\mathbf{T}^{\mathrm{smooth}}_{\mathrm{align},t}$ denotes the smoothed local-to-map alignment, and ${}^{B}\mathbf{T}_{C,t}$ represents the head-camera pose relative to the mobile base. The resulting ${}^{M}\mathbf{T}_{B,t}$ gives the base pose in the shared map frame. We extract its planar pose $(x,y,\psi)$, consisting of the map-frame position $(x,y)$ and yaw $\psi$, for use as policy input and in feedback control.

\subsection{Map-Aware Visuomotor Policies Learning}

MAVP uses the shared map to provide a common spatial reference for the base-related components of the policy state and action. Let $\mathbf s_t$ collect the visuomotor observations available at time $t$, and let $\mathbf b_t=(x_t,y_t,\psi_t)$ denote the current base pose in the shared map frame obtained from Eq.~\ref{eq:map-base-pose}. The policy $\pi_\theta$, parameterised by $\theta$, predicts a sequence of $H$ future actions, where $H$ is the action horizon:
\begin{equation}
\widehat{\mathbf A}_{t:t+H-1}
=
\pi_\theta(\mathbf s_t,\mathbf b_t),
\label{eq:policy-chunk}
\end{equation}
where $\widehat{\mathbf A}_{t:t+H-1}$ denotes the predicted action sequence from time $t$ to $t+H-1$. Action components depend on the task and robot interface; base motion is represented by target poses in the shared map frame. We refer to tracking these base-pose targets as \emph{map-frame pose control}.

The policy is trained by imitation learning from teleoperated demonstrations. Let
$\mathcal{D}=\{(\mathbf s_t,\mathbf b_t,\mathbf A_{t:t+H-1})\}$
denote the demonstration dataset. Imitation learning trains the policy to reproduce these demonstrated actions from the observed state by minimising
\begin{equation}
\mathcal{L}_{\mathrm{IL}}(\theta)
=
\mathbb{E}_{\mathcal D}
\left[
\ell\!\left(
\pi_\theta(\mathbf s_t,\mathbf b_t),
\mathbf A_{t:t+H-1}
\right)
\right],
\label{eq:imitation-learning}
\end{equation}
where the expectation is over samples from $\mathcal D$ and $\ell$ denotes the policy-specific imitation-learning loss. Different policy families instantiate $\ell$ differently, while sharing the same objective of matching the demonstrated behaviour.

During execution, the current base pose is obtained from visual localisation and is therefore subject to estimation noise. To expose the policy to similar variations during training, we first estimate the localisation-noise distribution from offline localisation runs. We use the empirical standard deviations of planar position and yaw fluctuations to define a zero-mean Gaussian with covariance
\[
\boldsymbol\Sigma
=
\operatorname{diag}(\sigma_x^2,\sigma_y^2,\sigma_\psi^2).
\]
Here, $\sigma_x$, $\sigma_y$, and $\sigma_\psi$ are the standard deviations of the noise in $x$, $y$, and yaw. For each training sample, we add an independent perturbation to the map-frame base pose:
\begin{equation}
\boldsymbol\epsilon_t
\sim
\mathcal N(\mathbf 0,\boldsymbol\Sigma),
\qquad
\widetilde{\mathbf b}_t
=
\mathbf b_t+\boldsymbol\epsilon_t.
\label{eq:map-pose-augmentation}
\end{equation}
Here, $\boldsymbol\epsilon_t$ denotes the sampled localisation perturbation and $\widetilde{\mathbf b}_t$ the perturbed base pose. The demonstrated action sequence is kept unchanged, while the base-pose input $\mathbf b_t$ is replaced by $\widetilde{\mathbf b}_t$. The augmented imitation-learning objective is therefore
\begin{equation}
\mathcal{L}_{\mathrm{IL}}^{\mathrm{aug}}(\theta)
=
\mathbb{E}_{\substack{\mathcal D\\
\boldsymbol\epsilon_t\sim\mathcal N(\mathbf 0,\boldsymbol\Sigma)}}
\left[
\ell\!\left(
\pi_\theta(\mathbf s_t,\widetilde{\mathbf b}_t),
\mathbf A_{t:t+H-1}
\right)
\right],
\label{eq:augmented-imitation-learning}
\end{equation}
This augmentation encourages the policy to produce consistent actions under small localisation errors.

\section{Empirical Evaluation}

We evaluate MAVP on six real-world mobile manipulation tasks. We test whether a shared spatial reference improves execution, which mapping, policy, and control components contribute, and whether ongoing map localisation improves repeated task execution. We organise the evaluation around the following questions:
\begin{enumerate}\setlength{\itemsep}{0pt}\setlength{\parsep}{0pt}\setlength{\partopsep}{0pt}
\item[\textbf{Q1:}] Does map-frame pose control improve task success compared with unanchored velocity and episode-initialised odometry pose control?
\item[\textbf{Q2:}] With the same base-pose input, how do map-frame pose targets compare with base velocities?
\item[\textbf{Q3:}] What is the contribution of explicitly providing the current map-frame base pose to the policy?
\item[\textbf{Q4:}] Does pose-noise augmentation improve tolerance to errors in the policy's pose input?
\item[\textbf{Q5:}] How does pose-error feedback affect task success and base-target tracking?

\item[\textbf{Q6:}] Which map-construction components support complete maps and reliable held-out localisation?
\item[\textbf{Q7:}] Does MAVP improve task success across different visuomotor policy families?
\item[\textbf{Q8:}] Does ongoing map localisation improve repeated task execution compared with initial alignment only?
\end{enumerate}

\subsection{Common Experimental Setup}

\firstlead{Physical setup:}

\begin{wrapfigure}{r}{0.48\linewidth}
    \centering
    \includegraphics[width=\linewidth]{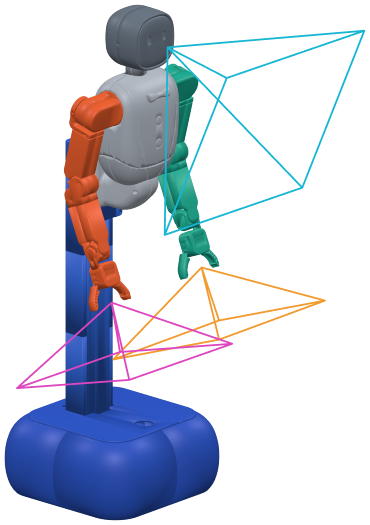}
    \caption{Configuration with one head-mounted and two wrist cameras.}
    \label{fig:robot-setup}
\end{wrapfigure}
The robot has a four-wheel omnidirectional base, a lifting and rotating
torso, and two 7-DoF arms with independent grippers. Base commands specify
planar velocity $(v_x,v_y)$ and yaw rate. Arm commands specify joint angles
and gripper states. The policy receives images from a head-mounted RGB camera
and two wrist RGB cameras. Figure~\ref{fig:robot-setup} shows the robot
configuration and camera placements.

Figure~\ref{fig:task-overview} shows Drawer Packing, Disassemble and Deliver, Lidded Box Packing, Conveyor Picking, Dual-Drawer Return, and Bag Packing in panels A to F. Insets provide close-up or additional views. Table~\ref{tab:task-definitions} gives the task sequences \mbox{and success criteria}.

\lead{Training setup:} We instantiate MAVP with three visuomotor policy families: ACT, DP, and FM. The controlled ablations in Q1--Q5 and the ongoing-localisation experiment in Q8 use ACT; Q7 compares all three families. We train all policies for 30{,}000 optimisation steps using automatic mixed precision and eight data-loader workers. ACT uses a batch size of 32 and a learning rate of $3\times10^{-5}$, with one observation step and an action chunk of 30 steps. The DiT-based implementations~\cite{peebles2023dit} used for Diffusion Policy (DP) and Flow Matching (FM) use a batch size of 64 and a learning rate of $2\times10^{-5}$, with two observation steps, a prediction horizon of 32 steps, and 24 action steps.

\lead{Evaluation metrics:} A demonstration or policy rollout succeeds when it completes every step of the corresponding task. Success rate is the percentage of successful rollouts. Each task has 50 teleoperated demonstrations, and each evaluated policy receives 50 \mbox{rollouts per task}.

\begin{table}[t]
\caption{Six mobile manipulation tasks; success requires completing every listed step.}
\label{tab:task-definitions}
\centering
\mavptablestyle
\begin{tabularx}{\linewidth}{@{}>{\raggedright\arraybackslash}p{0.29\linewidth}>{\raggedright\arraybackslash}X@{}}
\toprule
\textbf{Task} & \textbf{Description} \\
\midrule
Drawer Packing & Pick up the eraser, open the drawer, put the eraser and two toys inside, and close the drawer. \\
Disassemble and Deliver & Remove the peg, move to the box, and place all parts inside. \\
Conveyor Picking & Track and grasp an object on a moving conveyor. \\
Lidded Box Packing & Approach and open the box, place an object inside, and close the lid. \\
Bag Packing & Take a bag from the rack, put the table objects inside, and hang the bag back on the rack. \\
Dual-Drawer Return & Return the two objects outside the drawers to their corresponding drawers. \\
\bottomrule
\end{tabularx}
\end{table}

\begin{table}[t]
\centering
\caption{Success rates (\%) for the control variants. Pose denotes the map-frame base-pose input; aug. denotes training pose-noise augmentation.}
\label{tab:task-performance}
\mavptablestyle
\begin{tabular*}{\linewidth}{@{\extracolsep{\fill}}lcc@{}}
\toprule
\textbf{Method / variant} & \shortstack{\textbf{Disassemble}\\ \textbf{and Deliver}} & \shortstack{\textbf{Lidded Box}\\ \textbf{Packing}} \\
\midrule
Unanchored velocity & 38\% & 44\% \\
Odometry pose & 16\%  & 16\% \\
\midrule
Map-frame pose (w/o pose) & 74\% & 58\% \\
Velocity (w/ pose) & 60\% & 64\%\\
Map-frame pose (w/ pose) & 84\% & 76\%  \\
Map-frame pose (w/ pose + aug.) & \textbf{90\%} & \textbf{88\%} \\
\bottomrule
\end{tabular*}
\end{table}

\subsection{Map-Frame Pose Control (Q1, Q2, and Q3)}

\firstlead{Experimental setup:} Table~\ref{tab:task-performance} compares six visuomotor policy variants on Disassemble and Deliver and Lidded Box Packing. All variants receive arm joint and gripper state. In variant labels, pose denotes the current map-frame base pose and aug. denotes pose-noise augmentation during training.

\emph{Unanchored velocity} predicts base velocities without a map or base-pose input. \emph{Odometry pose} predicts pose targets in a frame initialised at the start of each episode. Its controller uses live odometry to execute those targets. \emph{Map-frame pose (w/ pose)} receives the current map-frame base pose and predicts targets in that frame.

\emph{Map-frame pose (w/o pose)} removes the base-pose input while retaining map-based localisation and base-pose targets. \emph{Velocity (w/ pose)} retains this input and predicts velocity commands. \emph{Map-frame pose (w/ pose + aug.)} adds the training augmentation from Eq.~\ref{eq:map-pose-augmentation}. This perturbs the pose input, preserving images, joint/gripper states, and action targets.

\begin{figure}[t]
\centering\begingroup
\setbox0=\hbox{\includegraphics[width=\linewidth]{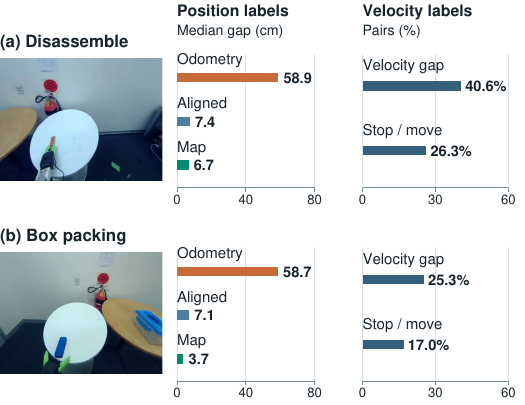}}
\setlength{\fboxsep}{0pt}
\makebox[0pt][l]{\usebox0}%
\makebox[0pt][l]{\raisebox{0.87\ht0}[0pt][0pt]{\colorbox{white}{\parbox[b]{0.30\linewidth}{\raggedright\sffamily\bfseries\fontsize{6.5}{7}\selectfont\strut (a) Disassemble\\and Deliver\strut}}}}%
\makebox[\linewidth][l]{\raisebox{0.38\ht0}[0pt][0pt]{\colorbox{white}{\parbox[b]{0.30\linewidth}{\raggedright\sffamily\bfseries\fontsize{6.5}{7}\selectfont\strut (b) Lidded Box\\Packing\strut}}}}%
\endgroup
\caption{Training-label variation for (a) Disassemble and Deliver and (b) Lidded Box Packing. Panels compare median position gaps across frames and fractions of nearby-state pairs with differing velocity or stop/move labels.}
\label{fig:position-velocity-diagnostics}
\end{figure}

Figure~\ref{fig:position-velocity-diagnostics} uses samples from the demonstration datasets that train the two task policies. The position-label plots report median pairwise position gaps among nearby demonstration states when expressed in three coordinate frames: raw odometry (episode-local), aligned odometry (odometry rigidly transformed into the shared map frame via $\mathbf{T}_{\mathrm{align}}$), and map coordinates (absolute poses from visual localisation). The gap measures label inconsistency across demonstrations. The velocity-label plots compare target velocities at nearby robot states. Together, they characterise training-label variation.

\begin{figure}[t]
\centering
\includegraphics[width=0.98\linewidth]{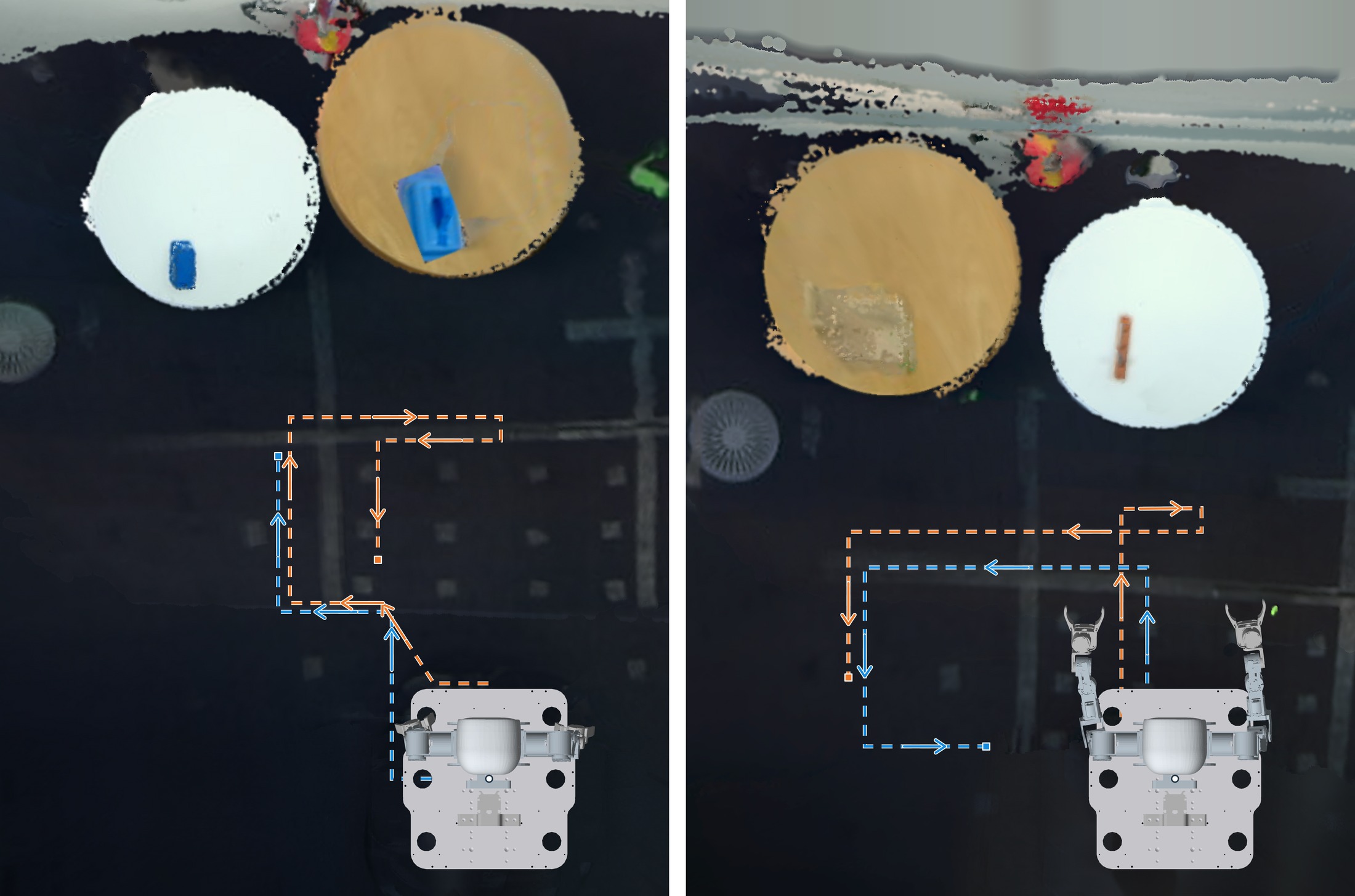}
\caption{Base trajectories for Lidded Box Packing (left) and Disassemble and Deliver (right). Velocity control (blue) stops early; map-frame pose control with training augmentation (yellow) continues toward task locations.}
\label{fig:task-closeups}
\end{figure}

\begin{figure}[t]
\centering
\includegraphics[width=0.95\linewidth]{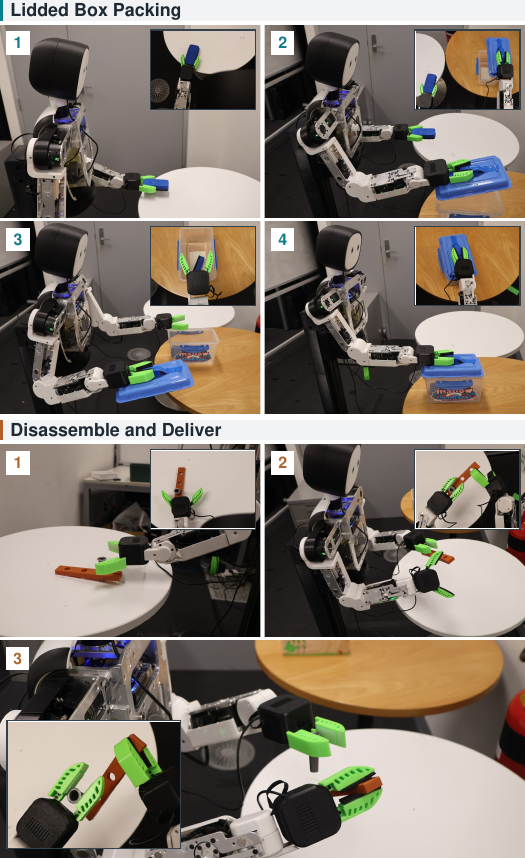}
\caption{Numbered execution snapshots for Lidded Box Packing and Disassemble and Deliver.}
\label{fig:task-sequences}
\end{figure}

\lead{Results:} Table~\ref{tab:task-performance} shows complementary benefits from map-frame targets, explicit pose inputs, and training augmentation. The complete configuration performs best on both tasks, while the controlled comparisons below distinguish the roles of its components. For Q1, map-frame pose control improves task completion over both unanchored velocity and episode-initialised odometry pose control. Its spatial targets give the controller an objective to track in the same frame used for demonstration supervision, so execution can respond to deviations from a target instead of relying only on predicted velocity commands. Figure~\ref{fig:task-closeups} contrasts velocity-controlled examples that stop short with augmented map-frame pose examples that continue towards manipulation locations; Figure~\ref{fig:task-sequences} shows task progression.

For Q2, \emph{Map-frame pose (w/ pose)} outperforms \emph{Velocity (w/ pose)} on both tasks without training augmentation. Both policies receive the current map-frame pose, so the difference concerns the action representation and its execution interface. A pose target specifies where the base should move, leaving the controller to determine the velocity needed to reach it. Velocity supervision also reflects demonstration timing, including acceleration, pauses, and the decision to stop; the nearby-state label variation in Fig.~\ref{fig:position-velocity-diagnostics} illustrates this dependence. This explains the advantage of spatial targets with feedback.

For Q3, the map-frame pose variant with pose input outperforms the variant without it, both trained without augmentation. The output representation is unchanged; the added input locates the base in the target frame, reducing the need to infer its location and heading from vision alone. 
\subsection{Robustness to Localisation Noise (Q4)}

\firstlead{Experimental setup:} Localisation errors affect the base pose supplied to the policy. Table~\ref{tab:localization-robustness} tests whether pose-noise augmentation improves tolerance to these errors. For each task, we compare map-frame pose policies trained with and without the augmentation in Eq.~\ref{eq:map-pose-augmentation}. At each policy inference, we add independent Gaussian noise to the pose input. The tested noise levels and their standard deviations $(\sigma_x=\sigma_y,\sigma_\psi)$ are no added noise $(0\,\mathrm{cm},0^\circ)$, low $(1\,\mathrm{cm},0.5^\circ)$, medium $(2\,\mathrm{cm},1^\circ)$, and high $(4\,\mathrm{cm},2^\circ)$. The controller uses the original localisation estimate. Entries report task success rates, with the no-added-noise results taken from Table~\ref{tab:task-performance}.

\begin{table}[t]
\centering
\caption{Success rates (\%) under Gaussian pose-input noise, with and without training augmentation.}
\label{tab:localization-robustness}
\mavptablestyle

\begin{adjustbox}{max width=\columnwidth}
\begin{tabular}{lcccc}
\toprule
\textbf{Policy variant}
& \textbf{No added noise}
& \textbf{Low}
& \textbf{Medium}
& \textbf{High} \\
\midrule
\multicolumn{5}{l}{\textit{Disassemble and Deliver}} \\
Map-frame pose (w/ pose)
& 84\% & 76\% & 68\% & 64\% \\
Map-frame pose (w/ pose + aug.)
& \textbf{90\%} & \textbf{88\%} & \textbf{78\%} & \textbf{68\%} \\
\midrule
\multicolumn{5}{l}{\textit{Lidded Box Packing}} \\
Map-frame pose (w/ pose)
& 76\% & 72\% & 68\% & 62\% \\
Map-frame pose (w/ pose + aug.)
& \textbf{88\%} & \textbf{90\%} & \textbf{86\%} & \textbf{76\%} \\
\bottomrule
\end{tabular}
\end{adjustbox}

\end{table}

\lead{Results:} Table~\ref{tab:localization-robustness} shows that pose-noise augmentation improves success at every tested perturbation level on both tasks. Increasing the perturbation generally makes execution less reliable, but augmentation retains an advantage despite variation between adjacent noise levels. The benefit is therefore not confined to the unperturbed evaluation condition.

This behaviour is consistent with the training objective: noisy pose inputs are paired with unchanged action targets, encouraging appropriate actions under pose noise.

\begin{figure}[t]
\centering
\includegraphics[width=\columnwidth]{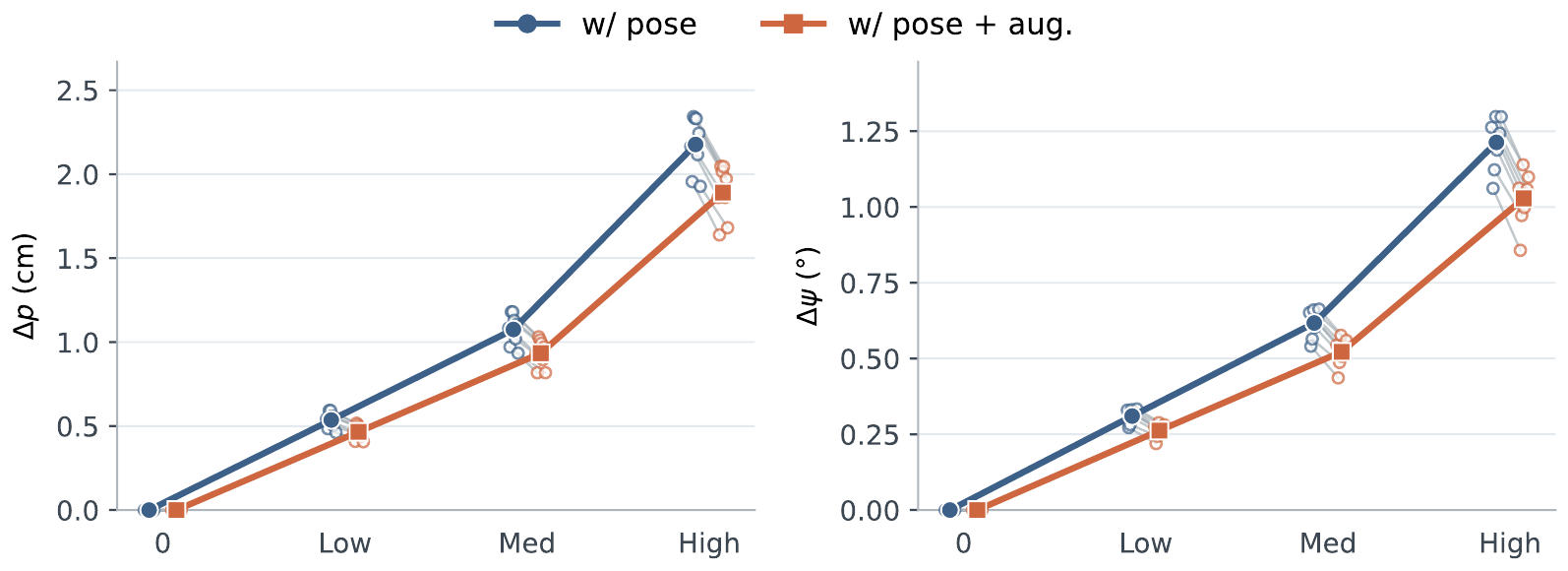}
\caption{Offline base-target changes under pose-input noise: $\Delta p$ is planar displacement and $\Delta\psi$ is wrapped yaw change. Open markers show recording means; solid curves weight recordings equally.}
\label{fig:pose-sensitivity}
\end{figure}

\lead{Pose-input sensitivity:} To complement Table~\ref{tab:localization-robustness}, we perturb the map-frame $(x,y,\psi)$ input offline at the same noise levels while holding reconstructed RGB and proprioceptive observations fixed. Both checkpoints receive identical observations and perturbation samples. We measure changes relative to each checkpoint's own unperturbed base-target prediction at a nominal $0.267$\,s horizon. Figure~\ref{fig:pose-sensitivity} shows smaller mean position and yaw changes for \emph{w/ pose + aug.} at every nonzero noise level, consistent with the task-success results.

\subsection{Contribution of Pose-Error Feedback (Q5)}

\begin{figure}[t]
\centering
\includegraphics[width=\linewidth]{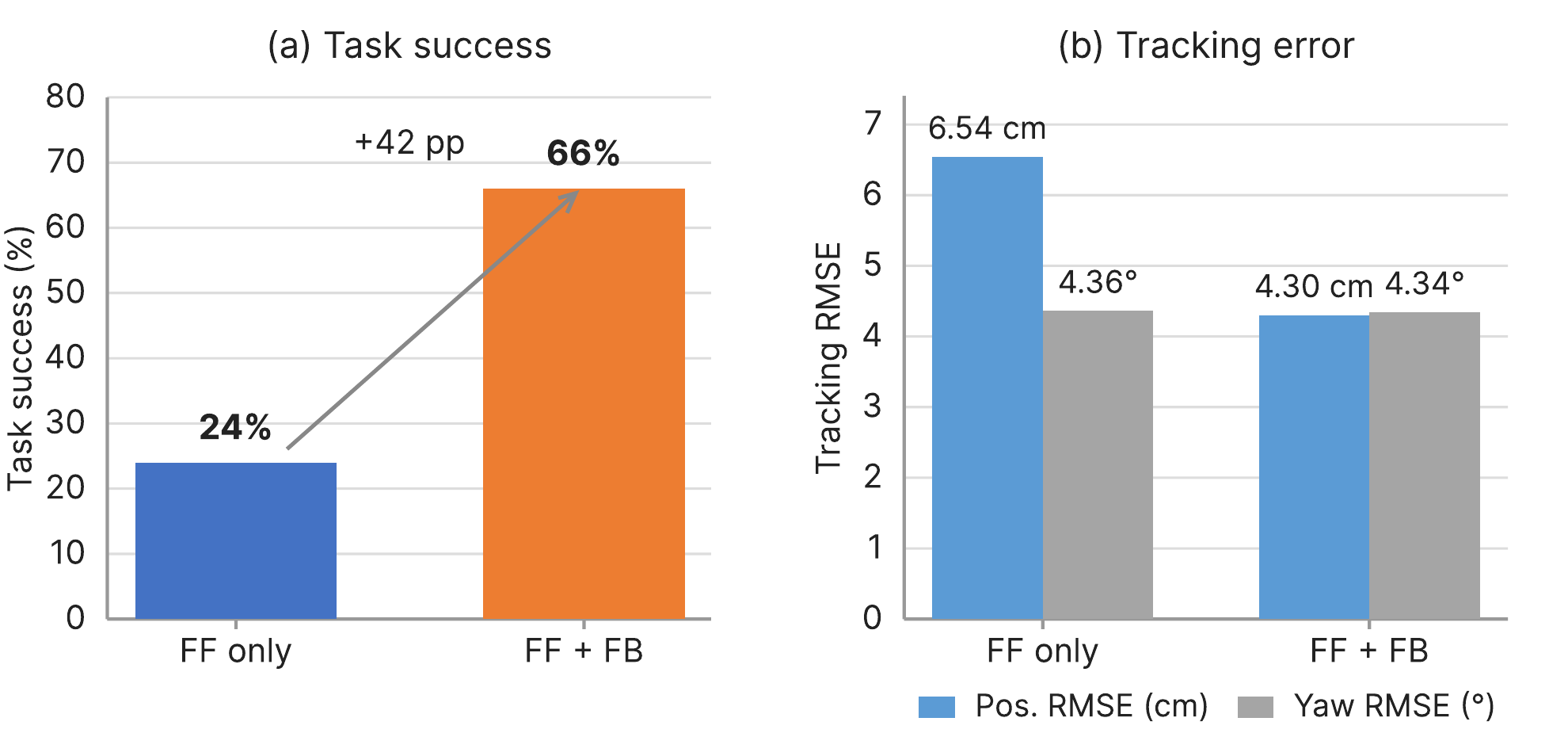}
\caption{Feedback ablation on Conveyor Picking: success and time-weighted tracking RMSE over 50 rollouts per variant using one augmented policy checkpoint. Feedback improves success and positional tracking, with little change in yaw RMSE.}
\label{fig:feedback-ablation}
\end{figure}

\firstlead{Experimental setup:} We compare FF only and FF + FB map-frame pose control on Conveyor Picking (Figure~\ref{fig:feedback-ablation}). Both use the same checkpoint trained with pose-noise augmentation; FF denotes feedforward and FB denotes pose-error feedback. FF only converts predicted pose-target changes into base velocities; FF + FB corrects the error between the target pose and current localisation estimate.

\lead{Metrics:} We report task success rate and time-weighted positional/yaw tracking RMSE (planar distance and wrapped yaw error between estimated base pose and active target, using localisation estimates).

\lead{Results:} Figure~\ref{fig:feedback-ablation} shows that adding pose-error feedback improves task success and positional tracking, while yaw tracking changes little. With the policy checkpoint fixed, this gain comes from target execution. Feedforward follows changes in those targets but does not respond to an offset between the target and the estimated base pose. Feedback corrects this offset to reach suitable picking positions.

\subsection{Geometric Map Construction Ablation (Q6)}

\firstlead{Shared protocol:}
Table~\ref{tab:map-consistency} evaluates the map-construction pipeline across the six manipulation tasks. For each task, the data used for map construction and localisation queries are disjoint. All variants use the same held-out stereo queries, reconstruction budget, calibration, feature extraction, and localisation thresholds. VGGT4D-based dynamic-region masking is retained in all variants. We ablate the remaining geometric components by removing robot-arm masking, restricting matching to temporal neighbours, removing temporal-neighbour matches, or replacing LightGlue with nearest-neighbour matching.

\lead{Metrics:}
A complete map is a single connected reconstruction that registers all selected reference images and passes the geometric quality checks. A stereo query is considered consistent when both views localise independently against the largest reconstructed component and their estimated poses agree within $3\,\mathrm{cm}$ and $2^\circ$. Deployable pairs additionally require the corresponding task map to be complete. Relative cost is the median ratio of feature-matching and reconstruction time to the full mapping pipeline. Stereo consistency measures agreement between the two views rather than absolute pose accuracy.

\lead{Results:}
The full mapping pipeline reconstructs complete maps for all six tasks. Temporal-only matching fails to produce a complete map, while the other ablations also reduce the number of deployable localisation queries. Some queries can still localise consistently within the largest component of an incomplete reconstruction, explaining why consistent-pair counts can remain high even when deployable-pair counts decrease.

\begin{table}[!t]
\caption{Geometric mapping results across the six manipulation tasks. Deployable pairs require a complete task map, and relative cost includes feature matching and reconstruction.}
\label{tab:map-consistency}
\centering
\mavptablestyle
\begin{tabularx}{\linewidth}{@{}>{\raggedright\arraybackslash}Xcccc@{}}
\toprule
\textbf{Variant} & \shortstack{\textbf{Complete}\\\textbf{maps}} & \shortstack{\textbf{Consistent}\\\textbf{pairs}} & \shortstack{\textbf{Deployable}\\\textbf{pairs}} & \shortstack{\textbf{Cost}\\\textbf{(rel.)}} \\
\midrule
Full mapping pipeline & \textbf{6/6} & \textbf{911/912} & \textbf{911/912} & 1.00 \\
Without arm mask & 3/6 & 908/912 & 600/912 & 1.58 \\
Temporal-only matching & 0/6 & 892/912 & 0/912 & \textbf{0.31} \\
Without temporal neighbours & 5/6 & 891/912 & 780/912 & 0.94 \\
Nearest-neighbour matching & 4/6 & 878/912 & 708/912 & 0.55 \\
\bottomrule
\end{tabularx}
\end{table}

\begin{table}[!t]
\centering
\caption{Success rates for ACT, DP, and FM over 50 rollouts per task. V: unanchored velocity without pose input; P: map-frame pose targets with pose input and localisation feedback.}
\label{tab:policy-family}
\mavptablestyle
\begin{tabularx}{\linewidth}{@{}>{\raggedright\arraybackslash}Xrrrrrr@{}}
\toprule
& \multicolumn{2}{c}{\textbf{ACT}}
& \multicolumn{2}{c}{\textbf{DP}}
& \multicolumn{2}{c@{}}{\textbf{FM}} \\
\cmidrule(lr){2-3}\cmidrule(lr){4-5}\cmidrule(l){6-7}
\textbf{Task} & V & P & V & P & V & P \\
\midrule
Disassemble and Deliver
& 38\% & \textbf{90\%}
& 42\% & 86\%
& 32\% & 88\% \\

Drawer Packing
& 20\% & 64\%
& 44\% & 62\%
& 38\% & \textbf{66\%} \\

Conveyor Picking
& 42\% & 66\%
& 38\% & 74\%
& 36\% & \textbf{76\%} \\

Lidded Box Packing
& 44\% & \textbf{88\%}
& 52\% & 56\%
& 50\% & 60\% \\

Bag Packing
& 18\% & 66\%
& 28\% & 68\%
& 32\% & \textbf{70\%} \\

Dual-Drawer Return
& 22\% & 68\%
& 30\% & 70\%
& 36\% & \textbf{74\%} \\
\bottomrule
\end{tabularx}
\end{table}

\subsection{Policy Architectures (Q7)}

\firstlead{Experimental setup:} Table~\ref{tab:policy-family} compares unanchored velocity control with map-frame pose control. Within each family, demonstrations, RGB and joint/gripper states, arm/gripper action representations, rollout counts, and success criteria are fixed. Unanchored velocity omits the map-frame base-pose input. Map-frame pose control includes it and tracks pose targets with feedback. ACT's two configurations match \emph{Unanchored velocity}/\emph{Map-frame pose (w/ pose + aug.)} in Table~\ref{tab:task-performance}. ACT uses action chunks~\cite{zhao2023aloha}, DP iterative denoising~\cite{chi2023diffusion}, and FM a learned vector field~\cite{lipman2022flow}.

\lead{Results:} Table~\ref{tab:policy-family} shows that map-frame pose control improves success across every evaluated task and policy family.  The common trend across ACT, DP, and FM supports the use of MAVP with different action-generation mechanisms. Each policy learns shared-map targets tracked using the current pose estimate, whether generated by a chunked transformer, iterative denoising, or a learned flow. This comparison extends the evidence across policy architectures, evaluating pose inputs, map-frame targets, and localisation feedback jointly; preceding ablations examine the individual choices.

\subsection{Ongoing Map Localisation (Q8)}
\label{sec:ongoing-map-localisation}

\firstlead{Experimental setup:} Table~\ref{tab:ongoing-map-localisation} compares \emph{initial alignment only} with \emph{ongoing map localisation} on Disassemble and Deliver and Lidded Box Packing. We reuse the same pose-noise-augmented map-frame pose checkpoint and FF + FB controller from the Q1--Q5 ablations; the ongoing-localisation numbers are taken from \emph{Map-frame pose (w/ pose + aug.)} in Table~\ref{tab:task-performance}. We evaluate 50 initial-only trials per task using the corresponding deployment settings and task-completion criteria.

\emph{Initial alignment only} fixes the local-to-map alignment after the first accepted map localisation in each deployment session and disables subsequent map queries. Local odometry and camera-to-base kinematics continue to update the pose supplied to both the policy and controller. Restarting the policy between trials retains this alignment and odometry state. \emph{Ongoing map localisation} instead updates the alignment using accepted map localisations during execution. Policy timing, action chunks, target lookahead, controller parameters, velocity limits, and task time limits follow the corresponding baseline settings, with no additional localisation noise. A local-tracking reset ends the initial-only session rather than triggering map-based recovery.

\begin{table}[t]
\centering
\caption{Task success (\%) with session-level initial alignment retained across policy restarts versus ongoing map localisation. Ongoing-localisation results are reused from 
Table~\ref{tab:task-performance}.}
\label{tab:ongoing-map-localisation}
\mavptablestyle
\begin{tabularx}{\linewidth}{@{}>{\raggedright\arraybackslash}Xcc@{}}
\toprule
\textbf{Task} & \shortstack{\textbf{Initial}\\\textbf{alignment only}} & \shortstack{\textbf{Ongoing map}\\\textbf{localisation}} \\
\midrule
Disassemble and Deliver & 64\% & \textbf{90}\% \\
Lidded Box Packing & 52\% & \textbf{88}\% \\
\bottomrule
\end{tabularx}
\end{table}

\lead{Results.} Table~\ref{tab:ongoing-map-localisation} shows that ongoing map localisation achieves higher task success on both tasks. This result indicates that continued localisation helps maintain a consistent spatial reference for the policy and controller. Accepted map updates correct the local-to-map alignment, keeping policy pose inputs and controller feedback referenced to the same map frame as the predicted targets.

\section{Conclusion and Future Work}
We presented MAVP, a map-aware visuomotor framework that gives mobile manipulation policies a shared spatial reference for learning and execution. MAVP reconstructs a map from teleoperated demonstrations, expresses demonstrated and predicted base poses in this common frame, and couples learned base-pose targets with localisation-based feedback control. This formulation separates \emph{where} the base should move from the low-level velocities required to reach that pose, while explicit pose conditioning and pose-noise augmentation improve robustness to spatial uncertainty. Across real-world tasks and multiple visuomotor policy families, the results consistently support shared-map pose targets over unanchored base control and demonstrate the value of feedback and continued localisation during execution. Future work will investigate more scalable map construction and localisation, richer uncertainty-aware policy and control mechanisms, and adaptation across diverse scenes and task configurations. Extending MAVP to larger workspaces, longer-horizon behaviours, and more general whole-body mobile manipulation offers a natural path towards reliable deployment in increasingly complex environments.

\bibliographystyle{IEEEtran}
\bibliography{reference}

\end{document}